%% file: main.tex
\documentclass{article}
\usepackage{amssymb}
\usepackage{times}
\usepackage{array}
\usepackage{booktabs}
\usepackage{graphicx}
\usepackage[font=small]{caption}
\usepackage{multirow}
\usepackage{placeins}
\usepackage{wrapfig} % Add to preamble
\usepackage{colortbl} % Required for coloring rows
\usepackage{xcolor}    % Extended color support
\usepackage{amsmath, amssymb, amsfonts}
\usepackage{pifont}
\usepackage[dvipsnames]{xcolor}

\usepackage[scale=1.1]{inconsolata} % nice monospaced font, slightly smaller

\usepackage{hyperref}
\definecolor{linkcolor}{HTML}{328DD3}
\hypersetup{
    colorlinks=true,
    linkcolor=black,
    citecolor=black,
    urlcolor=linkcolor
}

\newcommand{\greencheck}{{\color{Green}\ding{51}}}
\newcommand{\redcross}{{\color{red}\ding{55}}}

\RequirePackage[loading]{tracefnt}

\usepackage[numbers]{natbib}
\usepackage{multicol}
\usepackage{makecell}
\usepackage{float}
\usepackage[final]{corl_2025} % Uncomment for the camera-ready ``final'' version.

\title{Phantom: Training Robots Without Robots Using Only Human Videos}

\author{
    Marion Lepert$^{1}$, Jiaying Fang$^{1}$, Jeannette Bohg$^{1}$\\[0.1cm]
    $^{1}$Stanford University\\[2pt]
    \small\href{https://phantom-human-videos.github.io/}{%
        \ttfamily\textmd{https://phantom-human-videos.github.io/}%
    }\\[2pt]
    \vspace{-1cm}
}

\begin{document}
\maketitle

%===============================================================================
\vspace{-0.25cm}
\begin{abstract}
\input{01_abstract}
\end{abstract}

\keywords{Robot Imitation Learning, Learning from human videos}

\begin{figure}[h]
    \centering
    \includegraphics[width=\textwidth]{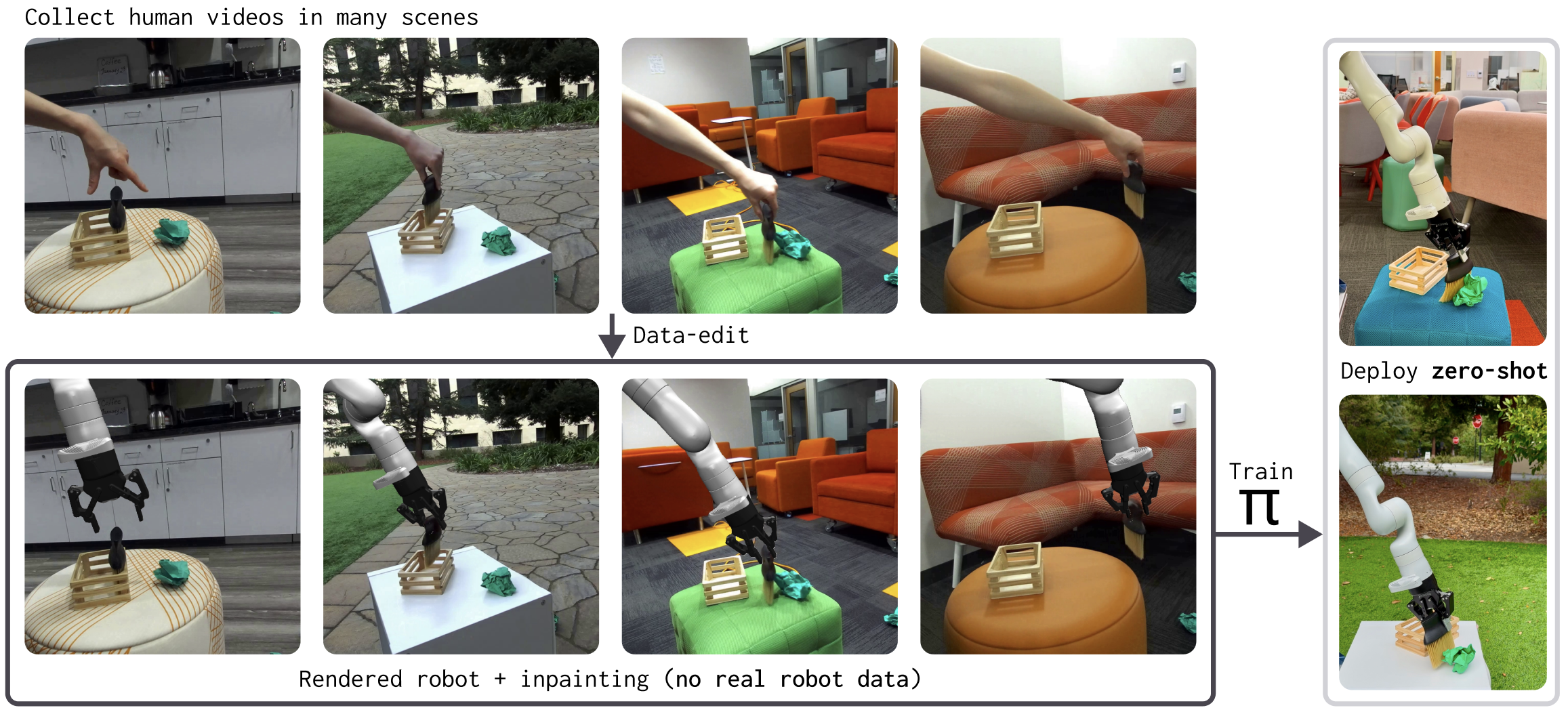}
    \caption{\textbf{Overview of Phantom.} Our method enables training robot policies without collecting any robot data. We first collect human video demonstrations in diverse environments and use inpainting to remove the human hand. A rendered robot is then inserted into the scene using the estimated hand pose. The resulting augmented dataset is used to train an imitation learning policy, which is deployed zero-shot on a real robot.}
    \vspace{-1em}
    \label{fig:main}
\end{figure}

%===============================================================================

\section{Introduction}
\input{02_intro}

\vspace{-0.5em}
\section{Related Works}
\input{03_related_works}

\vspace{-0.2em}

\section{Approach}
\input{04_approach}

\vspace{-0.2em}
\section{Results}
\input{05_results}

\vspace{-0.2em}
\section{Conclusion}
% \vspace{-1em}
\input{07_conclusion}

\clearpage
\section{Limitations}
\input{06_limitations}

%===============================================================================

\clearpage
% The acknowledgments are automatically included only in the final and preprint versions of the paper.
\acknowledgments{
This work was supported by the NSF through grant number \#2327974 as well as Intrinsic. We thank Jimmy Wu for help with hardware, and Claire Chen, Priya Sundaresan, Juntao Ren, and Zi-ang Cao for their valuable help throughout the project.}

%===============================================================================

\bibliography{references}  % .bib

\clearpage
\input{08_appendix}

\end{document}

%% file: 01_abstract.tex
Training general-purpose robots requires learning from large and diverse data sources. Current approaches rely heavily on teleoperated demonstrations which are difficult to scale. We present a scalable framework for training manipulation policies directly from human video demonstrations, requiring no robot data. Our method converts human demonstrations into robot-compatible observation-action pairs using hand pose estimation and visual data editing. We inpaint the human arm and overlay a rendered robot to align the visual domains. This enables zero-shot deployment on real hardware without any fine-tuning. We demonstrate strong success rates—up to 92\%—on a range of tasks including deformable object manipulation, multi-object sweeping, and insertion. Our approach generalizes to novel environments and supports closed-loop execution. By demonstrating that effective policies can be trained using only human videos, our method broadens the path to scalable robot learning. 

% Videos are available at \small\href{https://masquerade-robot.github.io}{\ttfamily\textmd{https://masquerade-robot.github.io}}

% {\ttfamily\url{https://phantom-human-videos.github.io/}}.

%% file: 02_intro.tex
Data scarcity remains a key challenge in advancing robotics research. While large-scale data collection efforts are gaining momentum, robotics datasets \cite{open_x_embodiment_rt_x_2023, pi2024, geminiroboticsteam2025geminiroboticsbringingai} remain orders of magnitude smaller than those used to train generalist vision and language models. These efforts are constrained by the slow and costly process of collecting data with robotics hardware. Moreover, increasing the quantity of data alone is not enough—\textit{diversity} in data is equally critical for generalization \cite{gao2024efficient}. Collecting diverse data across many environments with physical robots remains a formidable challenge.

Human videos offer a compelling alternative: they are abundant, diverse, and rich in task information. However, leveraging them for robot learning poses two key challenges: (1) human videos lack explicit action labels, and (2) human appearance differs substantially from robots. Prior methods trying to leverage human videos typically rely on co-training with robot data \cite{ren2025motion, bharadhwaj2025track2act, kareer2024egomimic, wang2023mimicplay, jain2024vid2robot} or reinforcement learning \cite{pertsch2022crossdomaintransfersemanticskill, xiong2021learning, chen2021learning, shao2020concept}. Such approaches fail to extract sufficient learning signals from human data alone, necessitating robot data to bridge the gap. These strategies are fundamentally limited in scalability. 

Our goal is to remove these bottlenecks by developing a framework that can scale with human videos alone. Our work focuses on the emerging data regime in which human videos become orders of magnitude more prevalent than robot demonstrations. While we do not demonstrate large-scale deployment in this work, we address the asymptotic setting where human data grows rapidly while robot data remains scarce. To that end, we propose an approach that requires only human video demonstrations to train a robot policy. Our method converts human videos into data-edited, {\em Phantom\/} robot demonstrations by extracting actions using a human hand pose estimator and replacing the human arm with a rendered robot. We then train a closed-loop imitation learning policy on these demonstrations and deploy it zero-shot on a real robot. Our method achieves high success rates across six tasks—including deformable object manipulation and generalization to novel environments—without requiring any robot data.

Our work draws inspiration from recent success in data-editing methods for robot-to-robot transfer \cite{chen2024mirage, chen2024roviaug, lepertshadow}. These techniques rely on precise proprioception and action labels, and until now, have not been effectively adapted to the more challenging human-to-robot setting. Our method does exactly this, leveraging data-editing for human-to-robot transfer. That such a method works and generalizes across diverse environments underscores a striking and perhaps unexpected insight: human demonstrations alone, when subject to data editing and careful hand pose estimation, can be directly leveraged for training a robot policy. Our method requires no robot data, no manual annotations, no object models, and can be deployed on any robot capable of executing the task.

\textbf{To summarize, our main contribution is demonstrating that data-editing-based cross-embodiment learning techniques are adaptable to human-to-robot transfer, which unlocks their utility for scaling robot learning through human video demonstrations.}

%% file: 03_related_works.tex
\subsection{Learning from Human Videos}
Many works have explored leveraging human videos to improve robot policies. A prominent line of research focuses on using diverse in-the-wild videos (e.g., YouTube) to improve generalization. Common strategies include pre-training visual representations \cite{kim2024openvla, radosavovic2023real, xie2024decomposing, nair2022r3m}, learning reward functions \cite{ma2022vip, pertsch2022crossdomaintransfersemanticskill, chen2021learning, shao2020concept}, building world models \cite{yang2023learning, du2023learning, liang2024dreamitate, bharadhwaj2024gen2act, cheang2024gr}, and learning object motion priors \cite{bharadhwaj2025track2act, bahl2023affordances, wang2024vlm}. These methods struggle to overcome the wide embodiment gap between humans and robots and rely extensively on robot data. \citet{shi2025zeromimicdistillingroboticmanipulation}
needs no robot data but only models post-grasp motion and requires manually collecting a goal image in the target scene.

Alternative approaches leverage curated human video demonstrations, which simplify the problem by ensuring that videos explicitly show task-relevant behaviors. While these videos must be manually collected, they are faster to gather relative to robot teleoperation data. Some works use human video demonstrations to learn motion priors \cite{bahl2022human,  wang2023mimicplay, xu2023xskill, bharadhwaj2024towards}. MimicPlay \cite{wang2023mimicplay} trained a high-level planner using human videos alongside a plan-guided imitation learning policy trained with robot demonstrations. Other methods \cite{jang2022bc, jain2024vid2robot} rely on paired human video and robot data to bridge the embodiment gap. 

Object-centric approaches have also been explored as an alternative direction \cite{hsu2024spot, heppert2024ditto, bahety2024screwmimic}. These methods typically estimate and track target object poses from video demonstrations, and learn a robot policy conditioned on the extracted object trajectories \cite{zhu2024vision}. However, they require identifying objects of interest and estimating rigid-body transformations which makes it hard to apply them to scenarios with deformables, granular materials, or multiple objects. Flow-based methods \cite{xiong2021learning, papagiannis2024r+, wen2023any, xu2024flow, ren2025motion} address some of these limitations by tracking trajectories of points instead of rigid body transformations. \citet{wen2023any, ren2025motion} track points on the human embodiment, which provides information on the general direction the robot should move in, but because humans and robots move in different ways, these methods still require robot data to refine the motion. Conversely, \citet{xu2024flow} only tracks flow on the manipulated object, but relies on object detection and simulation environments to refine robot motions. \cite{papagiannis2024r+} and \cite{haldar2025point} exclusively use human data but \cite{papagiannis2024r+} is limited to open-loop execution and \cite{haldar2025point} requires manual annotations of semantically meaningful object points. In contrast, our method is closed-loop, not bottlenecked by an object detector or the need for manual annotation, and works equally well on rigid, deformable, and multiple objects. A summary comparison of our method with prior work across key dimensions is provided in Table~\ref{table:related_works_comparison} in the appendix.

% TODO: tie back to the autoregressive generalsit policy argument 

\begin{figure*}
    \centering
    \includegraphics[width=\textwidth]{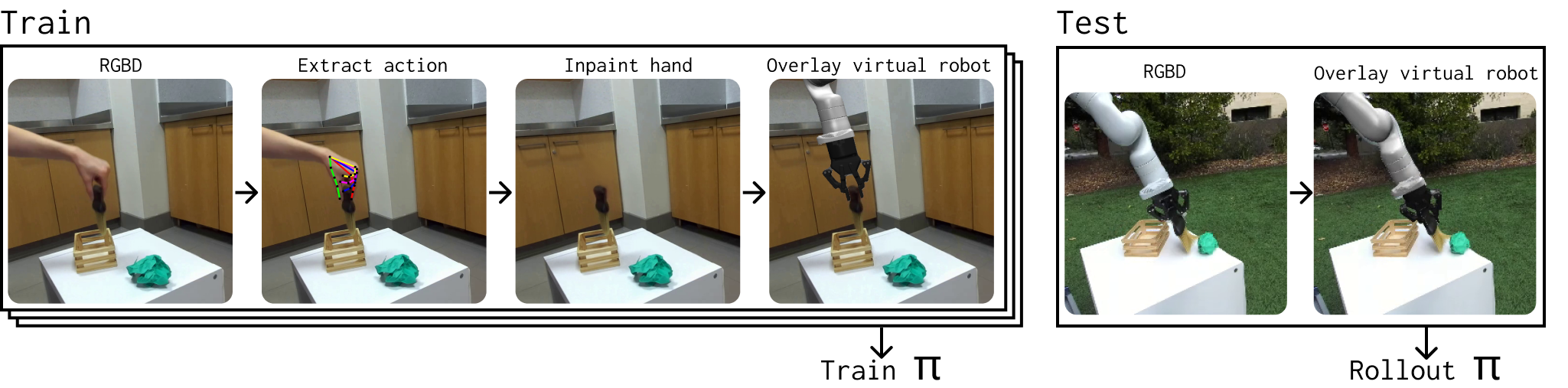}
    \caption{\textbf{Overview of our pipeline for learning robot policies from human videos.} During training, we first estimate the hand pose in each frame of a human video demonstration and convert it into a robot action. We then remove the human hand using inpainting and overlay a virtual robot in its place. The resulting augmented dataset is used to train an imitation learning policy, $\pi$. At test time, we overlay a virtual robot on real robot observations to ensure visual consistency, enabling direct deployment of the learned policy on a real robot.}
    \vspace{-1.2em}
    \label{fig:method}
\end{figure*}

\vspace{-0.3em}
\subsection{Data Editing for Cross-Embodiment Learning}
While many works focus on human-to-robot transfer, robot-to-robot cross-embodiment learning is also gaining attention as large-scale robotics datasets increasingly incorporate diverse data sources. Vision-based policies face significant challenges with cross-embodiment learning due to distribution shifts caused by the varying appearances of different embodiments. To address this, several methods propose data-editing strategies to mitigate these shifts. RoviAug \cite{chen2024roviaug} uses inpainting during training to remove the source embodiment from images and overlays a virtual rendering of the target embodiment in the same pose. At test time, the policy is deployed directly on the target embodiment. Shadow \cite{lepertshadow} replaces both the source and target robots with composite segmentation masks at train and test time, ensuring a close match between the input data distributions. Other methods, such as EgoMimic \cite{kareer2024egomimic} and AR2-D2 \cite{duan2023ar2}, adapt data-editing techniques for human-to-robot transfer. EgoMimic masks out each embodiment and overlays a red line along each arm, while AR2-D2 employs the same inpainting and virtual overlay strategy as \cite{chen2024roviaug}. However, both methods still rely on co-training with robot data to bridge the human-to-robot embodiment gap.

%% file: 04_approach.tex
\subsection{Problem Setup}
We assume access to a dataset $\mathcal{D}_{\text{human}} = \{\tau_h^i\}_{i=1}^N$ of $N$ human ($h$) video demonstrations $\tau_h^i$ of a manipulation task. Each demonstration consists of a sequence of images $\{I_{h,t}\}_{t=1}^{T}$ captured from a third-person viewpoint using an RGBD camera. The demonstration is performed using a pinch grasp with the thumb and index finger. 

Our goal is to use only these human video demonstrations to  train a closed-loop policy using imitation learning that can be deployed zero-shot on a target robot ($r$) for which no data has ever been collected. To do so, we use a data-editing strategy to convert our dataset $\mathcal{D}_{\text{human}}$ into $\mathcal{D}_{\text{robot}} = \{\tau_{r}^i\}_{i=1}^N$. Our objective is to convert each frame of a human demonstration into a corresponding robot observation-action pair: $I_{h,t} \rightarrow (I_{r,t}, a_{r,t})$. The goal of data-editing is for each $I_{r,t}$ to be drawn from the same distribution as images at test time on the target robot. Then, we can simply train our imitation learning policy on $\mathcal{D}_{\text{robot}}$ and deploy it on test-time robot observations.

Each robot action $a_{r,t}$ consists of the position and orientation of the end-effector and the opening width of the gripper $a_{r,t} = (\mathbf{p}_t, \mathbf{R}_t, g_t)$, where $\mathbf{p}_t \in \mathbb{R}^3$ is the Cartesian position of the end-effector, $\mathbf{R}_t \in \mathbb{R}^{6}$ represents the orientation of the end-effector using a 6D continuous rotation representation, and $g_t \in [0, 1]$ is the normalized opening width of the gripper.

We assume the deployment setup camera extrinsics are known. While train and test scenes do not need to match, we assume the camera used for data collection is at a similar viewpoint to that used at test time. This constraint could be relaxed by collecting data across a broader range of viewpoints, as in \cite{khazatsky2024droid}, but such large-scale data collection is beyond the scope of this work.

% While the scenes at train and test time do not need to match, we assume that the height of the camera used to collect videos is similar to that of the camera used to deploy the policy. Data collection is also assumed to take place with extrinsics similar to that of the test-time deployment setup. These requirements could be alleviated by collecting more data from a wide range of camera heights and collecting data in a wider area relative to the camera, as done in \cite{khazatsky2024droid}, but this amount of data collection is outside the scope of this paper. We also assume that the extrinsics of the camera used for test-time deployment are known.

% While the scenes at train and test time do not need to match, we assume that the height of the camera115
% used to collect videos is similar to that of the camera used to deploy the policy. This requirement116
% could be alleviated by collecting more data from a wide range of heights, as done in [43], but this117
% amount of data collection is outside the scope of this paper. We also assume that the extrinsics of118
% the camera used for test-time deployment are known.119

% We also assume that the extrinsics of the camera  used to deploy the policy on the robot are known.  - note that does not effect data collection itself, only affects data editing and deployment but not data collection. in support of htis ragument, please see appendix

\vspace{-0.6em}
\subsection{Action Labeling of Human Videos} \label{section:action_labels}
Since human videos lack explicit action information, we first address how to go from a frame of the human video $I_{h,t+1}$ to the corresponding robot action for the previous frame $a_{r,t} = (\mathbf{p}_t, \mathbf{R}_t, g_t)$. We start by estimating the human hand pose at each frame $I_{h,t}$ using HaMeR \cite{pavlakos2024reconstructing}. HaMeR predicts 21 keypoints, $\hat{\mathbf{X}}_t \in \mathbb{R}^{21 \times 3}$, corresponding to anatomical landmarks, along with a dense set of 778 vertices, $\hat{\mathbf{V}}_t \in \mathbb{R}^{778 \times 3}$, representing the hand mesh.

\begin{wrapfigure}{r}{0.5\textwidth}
    % \centering
    \includegraphics[width=\linewidth]{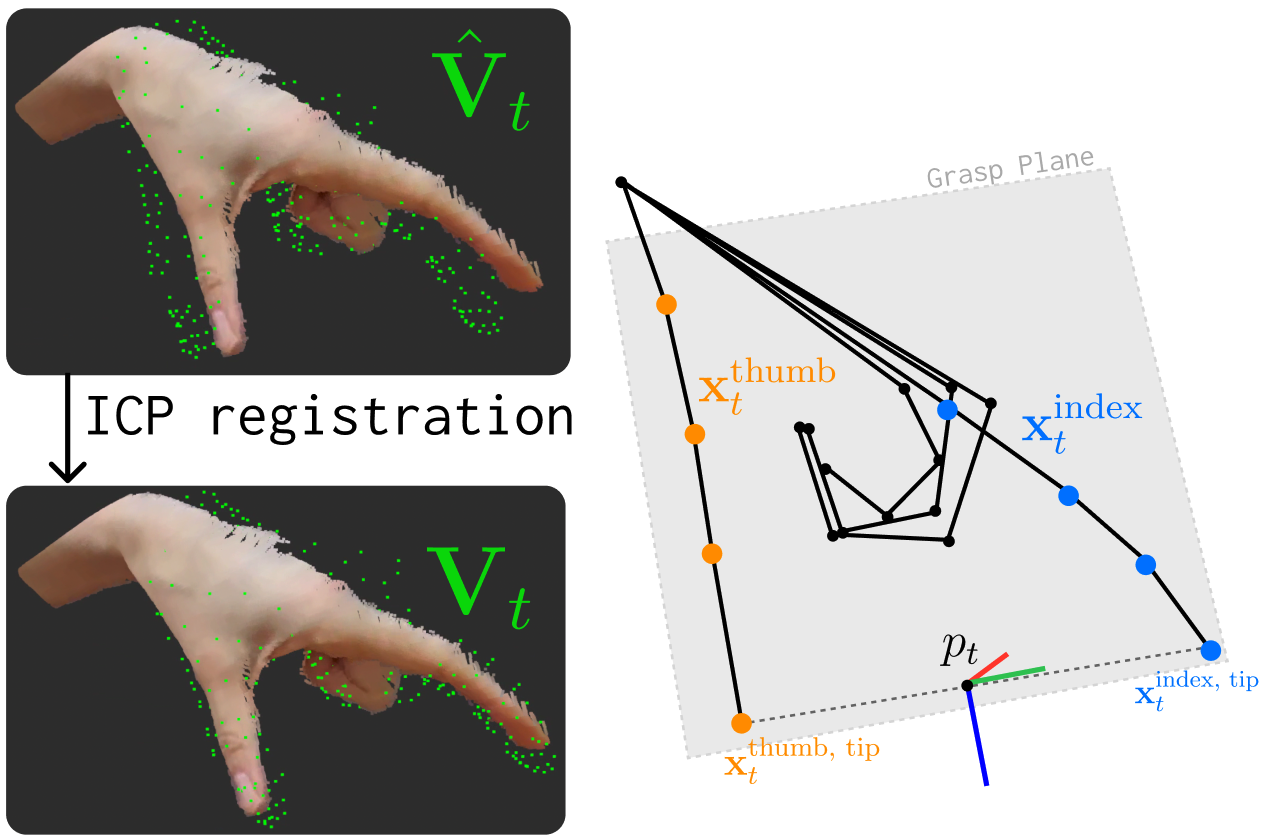}
    \caption{\textbf{Left}: To refine the HaMeR predicted mesh points $\hat{\mathbf{V}}_t$ shown in green, we use ICP registration to align them with the partial point cloud of the hand, $\mathbf{P}_t$ to obtain $\mathbf{V}_t$. \textbf{Right}: After aligning the HaMeR keypoints with the hand point cloud, we calculate the target position $\mathbf{p}_t$. }
    \label{fig:hand_keypoints}
    \vspace{-10pt}
\end{wrapfigure}

While HaMeR accurately captures hand shape, it struggles to estimate the absolute 3D pose due to its reliance on a monocular image. To refine this estimate, we incorporate depth data. We segment the hand in the RGB image using SAM2 \cite{ravi2024sam2}, yielding a segmentation mask $M_t$. Using $M_t$ and the corresponding depth image $D_t$, we extract a partial hand point cloud, $\mathbf{P}_t$. We then align the HaMeR-predicted mesh $\hat{\mathbf{V}}_t$ with $\mathbf{P}_t$ via Iterative Closest Point (ICP) registration, obtaining the optimal rigid transformation $\mathbf{T}_t \in SE(3)$ such that $\mathbf{P}_t \approx \mathbf{V}_t = \mathbf{T}_t \hat{\mathbf{V}}_t$ (see Fig. \ref{fig:hand_keypoints}). Since $\hat{\mathbf{V}}_t$ and $\hat{\mathbf{X}}_t$ are internally consistent, we apply $\mathbf{T}_t$ to the predicted keypoints to refine their positions: $\mathbf{X}_t = \mathbf{T}_t \hat{\mathbf{X}}_t$.

HaMeR also struggles with keypoints that are occluded in the RGB image—an issue exacerbated during grasping. Since HaMeR models all hand joints as ball joints, it often predicts unrealistic finger configurations under occlusion. To address this, we constrain the last two joints of the thumb and index fingers to a single degree of freedom, limiting their movement to anatomically feasible ranges. This ensures more accurate finger pose estimation when occlusions occur.

We use the refined keypoints $\mathbf{X}_t$ to define a target action for our policy, visualized in Fig. \ref{fig:hand_keypoints}. The target position, $\mathbf{p}_t$, is set as the midpoint between the keypoints at the tips of the thumb, $\mathbf{x}_t^{\text{thumb, tip}}$, and index finger, $\mathbf{x}_t^{\text{index, tip}}$. For the target orientation, $\mathbf{R}_t$, we fit a plane through all the keypoints of the thumb $\mathbf{x}_t^{\text{thumb}}$ and index finger $\mathbf{x}_t^{\text{index}}$ and compute a principal axis by fitting a vector through the keypoints of the thumb. $\mathbf{R}_t$ is then defined using the normal of this plane and the fitted vector. The gripper opening $g_t$ is computed as the distance between the keypoints corresponding to the fingertips of the thumb and index finger, $\mathbf{x}_t^{\text{thumb, tip}}$ and $\mathbf{x}_t^{\text{index, tip}}$. To mitigate slippage during grasping, we enforce a threshold, setting the bottom 20th percentile of predicted gripper distances in a single trajectory to fully closed.

HaMeR predicts keypoints in the camera's reference frame, meaning that $p_t$ and $\mathbf{R}_t$ are also expressed in this frame. We convert them into the robot's frame using the known camera extrinsics of our target setup to obtain the final robot action $a_{r,t}$.

% \begin{wrapfigure}{r}{0.5\textwidth}
%   \vspace{-23pt}
%   \includegraphics[width=\linewidth]{figs/kinova_tasks.png}
%   \caption{The out-of-distribution evaluation scenes used to evaluate the sweeping task on the Kinova robot.}
%   \label{fig:ood_scenes}
%   \vspace{-10pt}
% \end{wrapfigure}

\vspace{-0.2em}
\subsection{Bridging the Visual Observation Gap}
\label{section:inpaint_details}
Human arms and hands appear visually distinct from robot arms and grippers. A vision-based policy trained solely on human demonstrations struggles to generalize to a robot embodiment. To address this, we adapt the data-editing scheme from Rovi-Aug \cite{chen2024roviaug} to the human-to-robot transfer setting to compute $I_{h,t} \rightarrow I_{r,t}$. The edited images are used to train an imitation learning policy, which is then deployed on the target robot.

\textbf{Data-editing during training}: Each frame in the training dataset contains an image of a human arm performing a task. To replace the human embodiment with a robot, we first segment out the pixels corresponding to the human arm using SAM2 \cite{ravi2024sam2}, and then remove the segmented arm via inpainting using E2FGVI \cite{liCvpr22vInpainting}. Next, we render a virtual model of the target robot with its end effector in the corresponding pose, obtained from Section~\ref{section:action_labels} (i.e. its end effector pose at $(\mathbf{p}_t, \mathbf{R}_t, g_t)$). Given the known camera extrinsics, we synthesize an image of the robot from the appropriate viewpoint and overlay it onto the original image. To ensure realistic occlusions, we use depth data to determine which parts of the overlaid robot should be masked by objects in the environment. The final result is an image that closely resembles a real robot completing the task, as illustrated in Figure~\ref{fig:method}.

\textbf{Data-editing at inference time}: At inference time, each observation image contains a real robot arm. However, training images feature a rendered robot arm, which may have slight discrepancies in color and texture. To minimize domain shift, we overlay a virtual robot arm onto the real robot in each observation image, ensuring consistency between train and test distributions. An alternative approach, as proposed in Rovi-Aug, is to introduce color variations in the overlays during training to make the policy robust to these shifts. However, since this strategy has already been explored in prior work, we opt for the simpler inference-time overlay approach.

%% file: 05_results.tex
We evaluate our method across a range of tasks that highlight the versatility of our method.  To demonstrate that our method works across different robots, we present results on both a Franka and a Kinova robot.  Policies are trained using Diffusion Policy \cite{chi2023diffusion}, with OSC and IK controllers used for low-level control of the Franka and Kinova, respectively. Virtual robot renderings are generated using Mujoco \cite{todorov2012mujoco}.

\subsection{Comparison of Data Editing Methods}
\begin{figure}[h]
  \centering
  % \vspace{-1em}
  \includegraphics[width=0.8\linewidth]{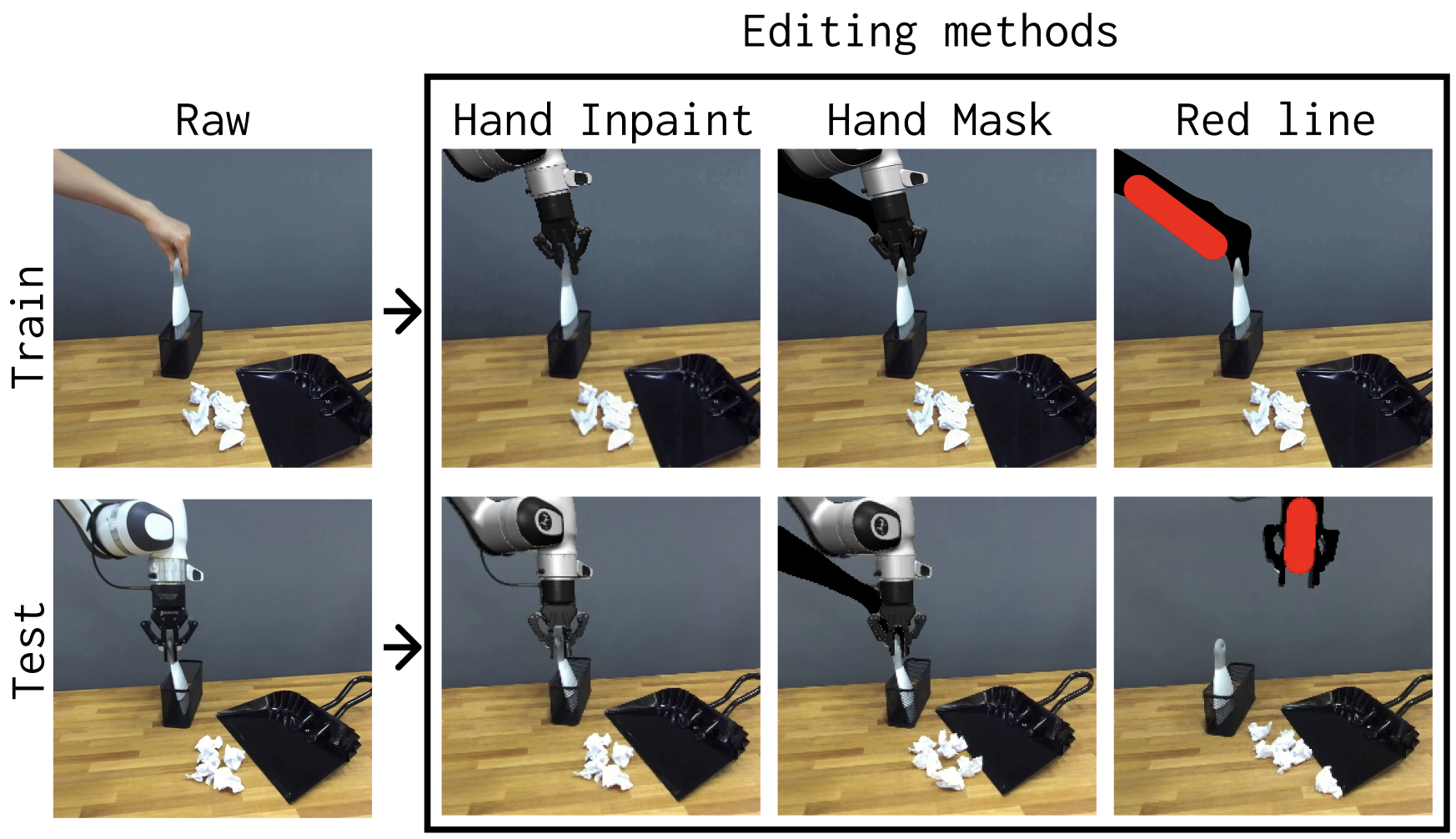}
  \caption{The three data-editing strategies we compare.}
  \label{fig:baselines}
\end{figure}

To the best of our knowledge, no prior (non-concurrent) published  work has trained closed-loop imitation learning policies using only human video demonstrations that can manipulate rigid and deformable objects, and groups of objects without requiring any manual annotations. %Therefore, we focus our experiments on identifying the most effective data-editing strategy for human-to-robot policy transfer. %(see detailed justification for the following baselines in Appendix \ref{section:baseline-justification}). 
After careful analysis of all candidate baselines (see detailed justification in Appendix \ref{section:baseline-justification}), we focus our experimental analysis on determining the most effective data-editing strategy for human-to-robot policy transfer (see Figure~\ref{fig:baselines}):

\textbf{Hand Inpaint (Phantom - Ours)}:  We adapt the data-editing strategy from Rovi-Aug \cite{chen2024roviaug}, which was developed for the simpler robot-to-robot setting, to the human-to-robot setting. During training, the human arm is segmented out and replaced with inpainting. An image of the target robot is synthesized using a virtual model from the appropriate viewpoint and overlaid onto the original image. At test time, a rendered robot arm is overlaid onto the real robot arm to minimize domain shift. See Section ~\ref{section:inpaint_details} for more details.

\textbf{Hand Mask}: We adapt the data-editing strategy from Shadow \cite{lepertshadow}. This method was also developed for the simpler robot-to-robot setting. During training, the human arm is masked out, and a virtual robot in the same pose is overlaid. While the original method overlays a black mask of the robot, we use an RGB image for a cleaner comparison with Hand Inpaint. At test time, a hand mask generated by a trained diffusion model is applied, and a virtual robot is overlaid on the real robot. See Appendix \ref{section:shadow_details} for more details.

% \begin{wrapfigure}{r}{0.6\textwidth}
%   \centering
%   \vspace{-15pt}
%   \includegraphics[width=\linewidth]{figs/baselines.png}
%   \caption{The three inpainting strategies we compare.}
%   \label{fig:baselines}
%   % \vspace{-10pt}
% \end{wrapfigure}

\textbf{Red Line}: EgoMimic \cite{kareer2024egomimic}  proposes a data-editing approach for learning robot policies from egocentric human videos. While we do not directly compare with their full method, as it requires robot data, we evaluate their data-editing strategy. The human arm is masked out in black during training and overlaid with a red line along its length. At test time, the robot arm is similarly blacked out, with a red line overlaid in the same manner. 

\textbf{Vanilla}: We also compare to a baseline that does not modify the train or test images in any way.

% \vspace{1em}
% \vspace{-2em}
\subsection{In-distribution Scene} 
\vspace{-0.2em}
We start by evaluating how well our method can transfer a policy trained exclusively on data-edited human video demonstrations collected in a single scene to a robot in the same scene. This evaluates how well our method bridges the physical and visual embodiment gap between human and robot without the added complexity of testing scene generalization. We evaluate our method on five tasks that highlight the diversity of skills our method can learn. For each task we collect between 250-350 human video demonstrations (see Appendix \ref{section:appendix-dataset}). 

Hand Inpaint and Hand Mask achieve high success rates across all tasks. However, Hand Mask takes on average 73\% longer to rollout due to having to run an additional diffusion model at test time to generate the hand masks. The Red Line data-editing strategy fails to complete any tasks, indicating that it does not adequately bridge the visual embodiment gap between humans and robots.

\vspace{-0.5em}
\begin{table}[H]
\centering
\setlength{\tabcolsep}{3pt}
\begin{tabular}{@{}l@{\hskip 12pt}c@{\hskip 8pt}c@{\hskip 8pt}c@{\hskip 8pt}c@{\hskip 10pt}|@{\hskip 10pt}c@{\hskip 8pt}c@{\hskip 8pt}c@{\hskip 8pt}c@{}}
\toprule
& \makecell{Pick/\\Place\\Book} & \makecell{Stack\\Cups} & \makecell{Tie\\Rope} & \makecell{Rotate\\Box} 
& \makecell{Grasp\\Brush} & \makecell{Sweep\\$> 0$} & \makecell{Sweep\\$> 2$} & \makecell{Sweep\\$> 4$} \\
\midrule
Hand Inpaint (Phantom) & 0.92  & 0.72 & 0.64 & 0.72 & 0.88  & 0.80 &  0.72 & 0.40 \\
Hand Mask              & 0.92  & 0.52 & 0.60 & 0.76 & 0.75  & 0.75 &  0.72 & 0.68 \\
Red Line               & 0.0   & 0.0  & 0.0  & 0.0  & 0.0   & 0.0  &  0.0  & 0.0 \\
Vanilla                & 0.0   & 0.0  & 0.0  & 0.0  & 0.0   & 0.0  &  0.0  & 0.0 \\
\bottomrule
\end{tabular}
\vspace{0.5em}
\caption{\textbf{In-distribution scene results}: Both Hand Inpaint and Hand Mask achieve high success rates across all tasks. The Red Line strategy fails to achieve success on any task, as does the Vanilla baseline. For the sweep task, we evaluate success at multiple levels of completion: Grasp Brush measures whether the robot successfully picks up the brush, while Sweep $> 0$, Sweep $> 2$, and Sweep $> 4$ indicate the number of pieces swept into the dustpan. 25 rollouts per evaluation.}
% \vspace{-1.5cm}
\label{table:merged_ablation}
\end{table}

\begin{figure}[h]
    \centering
    \includegraphics[width=\linewidth]{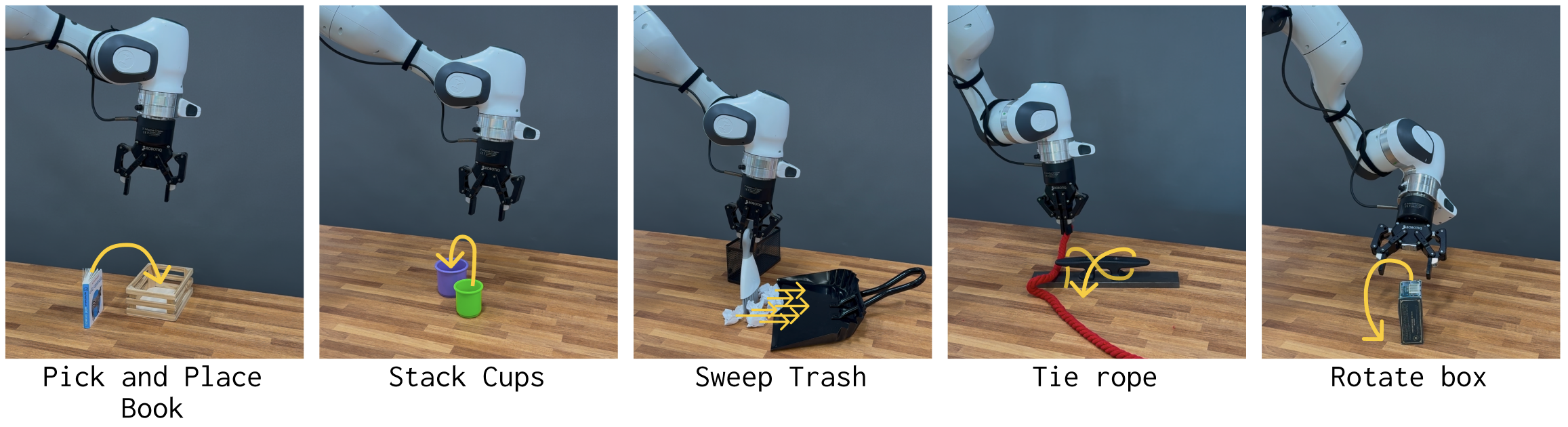}
    \caption{\textbf{Tasks evaluated in an in-distribution scene | Pick and Place Book}: The robot must pick up a book and place it inside a wooden container. \textbf{Stack Cups}: The robot must stack the green cup inside the purple cup. Precise alignment is critical, as the cups differ in diameter by only 1.5 cm. \textbf{Sweep Trash}: The robot must pick up a sweeper and sweep six pieces of trash into a dustpan. This task involves coordinated multi-object manipulation, requiring the robot to control the sweeper while simultaneously managing the movement of multiple loose objects. Additionally, the pieces of trash exhibit unpredictable dynamics, necessitating continuous adaptation based on real-time feedback. \textbf{Tie Rope}: The robot must tie a simplified cleat hitch, a sailing knot that follows a figure-eight $\infty$ pattern. This task is challenging due to the precise manipulation required of a highly deformable object. \textbf{Rotate Box}: The robot must rotate a box 90 degrees onto a new face in a controlled fashion (simply knocking it over is not valid).}
    \vspace{-1.5em}
    \label{fig:franka_tasks}
    % \vspace{-0.5cm}
\end{figure}
% \vspace{-1em}

\vspace{-2em}
\subsection{Out-of-distribution Scenes} 
\vspace{-0.5em}
Next, we evaluate how well our method generalizes to new, unseen environments. We collect human video demonstrations of a sweeping task across diverse scenes (see Fig. \ref{fig:main} for examples). The robot must grasp the sweeper and sweep the green piece of trash off the surface. We collect 950 human video demonstrations across a wide range of indoor and outdoor scenes (see Appendix \ref{section:appendix-dataset}). %(TODO: check numbers (see appendix)).

\begin{wrapfigure}{r}{0.4\textwidth}
  \vspace{-15pt}
  \includegraphics[width=\linewidth]{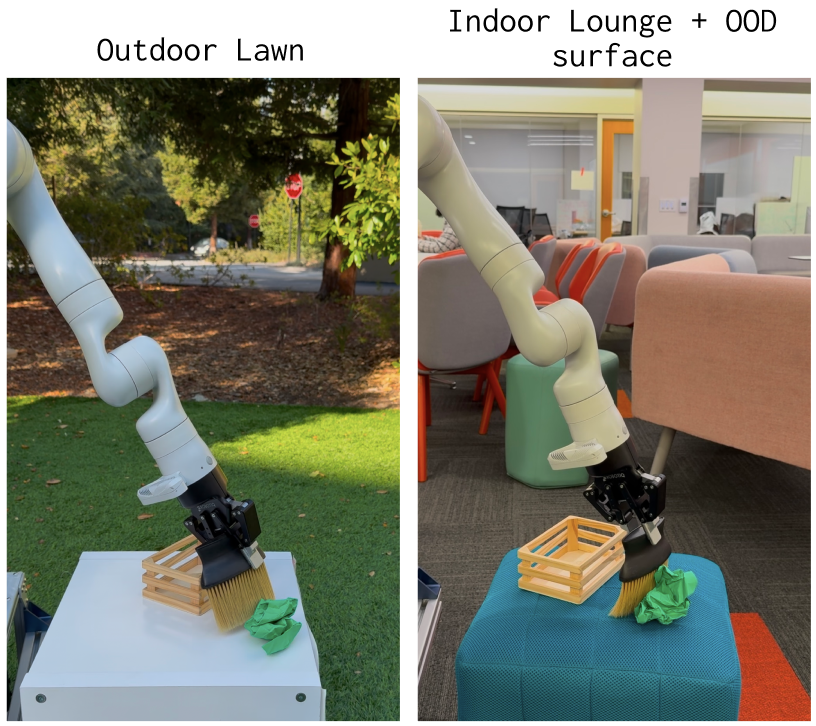}
  \caption{The out-of-distribution evaluation scenes used to evaluate the sweeping task on the Kinova robot.}
  \label{fig:ood_scenes}
  \vspace{-20pt}
\end{wrapfigure}

We assess generalization in three out-of-distribution (OOD) scenarios (see Fig \ref{fig:ood_scenes}) that were never seen during data collection. These consist of two OOD scenes — Outdoor Lawn and Indoor Lounge — and one OOD surface. Rollouts take place with dynamic background variations, including moving cars and passersby.

Hand Inpaint achieves high success rates across all three OOD environments. Its best performance is in the indoor lounge, which aligns with expectations, as 80\% of the training data was collected indoors. When evaluated on an unseen surface, performance drops by 20\%, likely due to the limited diversity of training surfaces (four, see Appendix \ref{section:appendix-dataset}.)

Overall, Hand Inpaint and Hand Mask perform comparably across both in-distribution scenes and out-of-distribution scenes.  Red Line and the Vanilla baseline fail on all tasks due to large visual mismatches between human and robot embodiments. While performance is comparable between Hand Inpaint and Hand Mask, Hand Inpaint is on average 73\% faster to rollout, as it does not require hand mask generation with a diffusion model, and produces training images that more closely resemble real-world robot data, making it the more effective solution.

\vspace{-1em}
\begin{table}[H]
\centering
\begin{tabular}{lccc}
\toprule
& \makecell{Outdoor lawn} & \makecell{Indoor lounge} &\makecell{Indoor lounge \\ + OOD surface} \\
\midrule
Hand Inpaint (Phantom) & 0.72 & 0.84 & 0.64\\
Hand Mask  & 0.52 & 0.76 & 0.68\\
\bottomrule
\end{tabular}
\vspace{0.3em}
\caption{\textbf{Out-of-distribution (OOD) scene results:} Success rates of policies trained on human video demos and tested in three unseen environments: Outdoor Lawn, Indoor Lounge, and Indoor Lounge with an OOD Surface. Hand Inpaint achieves the highest success rates across all settings. Hand Mask performs comparably but is worse in the outdoor lawn setting, perhaps due to weather variations during rollouts. 25 rollouts per eval.}
\label{table:ablation}
\end{table}

\vspace{-1cm}
 
% \vspace{1em}
\subsection{Evaluating the Need for High-quality In-painting}

\vspace{-0.2em}
\begin{figure}[h]
    \centering
    \includegraphics[width=0.8\linewidth]{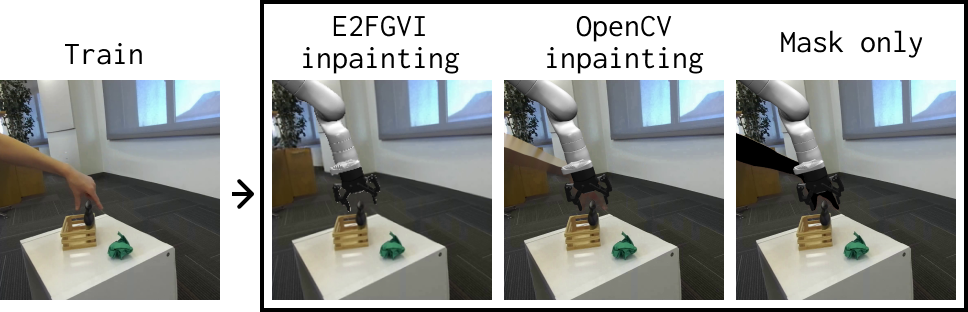}
    \caption{The three inpainting strategies we compare.}
    \label{fig:inpaint_quality}
\end{figure}
\vspace{-0.4em}

% \begin{figure}[h]
% \centering
% \begin{minipage}[t]{0.65\linewidth}
% \centering
% \includegraphics[width=\linewidth]{figs/inpaint_quality.png}
% \caption{The three inpainting strategies we compare.}
% \label{fig:inpaint_quality}
% \end{minipage}%
% \hspace{0.01\linewidth}
% \begin{minipage}[t]{0.25\linewidth}
% \centering
% % \vspace{0.1cm}
% % \begin{tabular}{lcc}
% \vspace{-8em}
% \begin{tabular}{l@{\hskip -3pt}c@{\hskip 4pt}c}
% \toprule
% & \makecell{Indoor Lounge} \\
% \midrule
% E2FGVI & 0.84 \\
% OpenCV & 0.76  \\
% Mask only & 0.60\\
% \bottomrule
% \end{tabular}
% \vspace{0.1cm}
% \captionof{table}{Comparison of in-painting methods. 25 rollouts per eval. }
% \label{table:ablation}
% \end{minipage}
% \end{figure}

At train time, our preferred Hand Inpaint method relies on inpainting to remove the human arm from the human video and replace it with a realistic background. We test how important this inpainting is by comparing three variations (see Fig. \ref{fig:inpaint_quality}) on the sweeping task in the unseen Indoor Lounge scene: (1) High-quality inpainting (E2FGVI \cite{liCvpr22vInpainting}): A state-of-the-art video inpainting method, (2) Low-quality inpainting (OpenCV inpaint): The OpenCV inpainting function removes the arm but leaves visible artifacts, (3) No-inpainting (Mask Only): The human arm is simply masked out during training. Unlike Hand Mask, no hand mask is overlaid at test time.

\begin{wraptable}{r}{0.4\textwidth}
\vspace{-10pt} % optional
\centering
\begin{tabular}{lcc}
\toprule
& \makecell{Indoor Lounge} \\
\midrule
E2FGVI inpaint & 0.84  \\
OpenCV inpaint & 0.76  \\
Mask only & 0.60\\
\bottomrule
\end{tabular}
\vspace{0.1cm}
\caption{Comparison of in-painting methods}
\label{table:ablation}
\vspace{-10pt} % optional
\end{wraptable}

High-quality inpainting with E2FGVI yields the best performance, achieving a 84\% success rate. However, low-quality inpainting performs surprisingly well, with a 76\% success rate. This suggests that there is enough variation in the low-quality inpainting at train time for the model to become agnostic to the artifacts. In contrast, using no-inpainting noticeably degrades performance. The mask-only approach results in a 24 percentage point performance drop.  These results suggest that while high-quality inpainting is ideal, our method still performs well with primitive inpainting. Additionally, the high performance of the mask-only approach relative to the 0\% success rate of the Red Line method in the easier in-distribution experiments implies that including an overlay of the robot at training time is essential. 

% \begin{table}[h]
% \centering
% \begin{tabular}{lcc}
% \toprule
% & \makecell{Indoor Lounge} \\
% \midrule
% \rowcolor{yellow!30} E2FGVI inpaint & 0.84  \\
% OpenCV inpaint & 0.76  \\
% Mask only & 0.60\\
% \bottomrule
% \end{tabular}
% \vspace{0.1cm}
% \caption{\textbf{Comparison of in-painting methods}: Using the highest quality inpainting method E2FGVI \cite{liCvpr22vInpainting} achieves the highest success rate, but the very primitive OpenCV inpainting function also does remarkably well. Using no inpainting at all leads to a 24 percentage point drop in performance. 25 rollouts per evaluation.}
% \label{table:ablation}
% \end{table}

% \begin{table}[h]
% \centering
% \begin{tabular}{lcc}
% \toprule
% & \makecell{Indoor Lounge} \\
% \midrule
% \rowcolor{yellow!30} E2FGVI inpaint & 0.84  \\
% OpenCV inpaint & 0.76  \\
% Mask only & 0.60\\
% \bottomrule
% \end{tabular}
% \vspace{0.1cm}
% \caption{Comparison of in-painting methods}
% \label{table:ablation}
% \end{table}

%% file: 07_conclusion.tex
We present a method for training robot policies without collecting any robot data, using only human video demonstrations. Our approach successfully transfers policies across a diverse set of tasks, including deformable object manipulation and multiple objects manipulation. Furthermore, we demonstrate zero-shot deployment in novel scenes, showing that human video demonstrations and robot rollouts do not need to occur in the same environment. This flexibility makes our method highly scalable and accessible. By enabling anyone with an RGBD camera to collect meaningful training data anywhere, we lower the barrier to large-scale robot learning and broaden the potential for real-world deployment.

Lastly, a promising direction in robotics is training autoregressive generalist policies on large-scale robotics datasets \cite{open_x_embodiment_rt_x_2023,kim2024openvla,pi2024}. Other interesting methods using human video demonstrations introduce complexities that are not amenable to such architectures \cite{xu2024flow,wang2023mimicplay,bahl2022human,zhu2024vision}. In contrast, our simple data-editing approach generates observation-action pairs of robots performing tasks, which can be easily integrated into datasets used to train these generalist policies — a promising avenue for future work.

%% file: 06_limitations.tex
% \begin{itemize}
%     \item The performance of our approach is limited by the performance of existing hand pose estimators since it relies on them to obtain the target actions from a demonstration video. Because hand pose estimators currently still struggle with occlusions, our method does too. However, this also means that our method will get better with time as hand pose estimators improve. 
%     \item Our method only works when the robot can follow the same strategy as the human to complete the task. As a result, our policy may lead the robot to  collide with the environment even though the human hand does not. Additionally, differences in the surface properties of a human fingertip and robot gripper may lead to different object motions. 

%     \item We limit our demonstrations to pinch grasps because our robots are limited to using parallel jaw grippers - a limitation that is shared with virtually all large-scale robot data collection efforts  \cite{khazatsky2024droid,open_x_embodiment_rt_x_2023,pi2024}.

%     \item We only assess quasi-static tasks, as we do not address the latency mismatch between a human demonstration and a trained policy rolled out on real hardware.

% \end{itemize}

Our approach has several limitations. First, its performance is limited by the performance of existing hand pose estimators since it relies on them to obtain the target actions from a demonstration video. Because hand pose estimators currently still struggle with occlusions, our method does too. However, this also means that our method will get better as hand pose estimators improve. Second, our method only works when the robot can follow the same strategy as the human to complete the task. As a result, our policy may lead the robot to collide with the environment even though the human hand does not. Additionally, differences in the surface properties of a human fingertip and robot gripper may lead to different object motions. Third, we limit our demonstrations to pinch grasps because our robots are limited to using parallel jaw grippers - a limitation that is shared with virtually all large-scale robot data collection efforts  \cite{khazatsky2024droid,open_x_embodiment_rt_x_2023,pi2024}. Fourth, we only assess quasi-static tasks, as we do not address the latency mismatch between a human demonstration and a trained policy rolled out on real hardware.

%% file: 08_appendix.tex
\section*{Appendix}

\subsection{Detailed justification for our baseline selection}
\label{section:baseline-justification}

In this section, we discuss the most relevant prior works and explain why we do not include direct comparisons to them in our experiments (with the exception of a partial comparison to EgoMimic). Table \ref{table:related_works_comparison} summarizes our detailed comparison between Phantom and related works. Broadly, we identify two key reasons why existing methods are not suitable baselines for our approach:

\begin{enumerate}
    \item \textbf{The robot data bottleneck}: Phantom is designed to explore how to build a highly scalable framework for learning from human videos. In contrast, many prior methods rely on co-training with teleoperated robot demonstrations (see Table. \ref{table:related_works_comparison}), which introduces a fundamental bottleneck to scalability. Collecting robot data is expensive, time-consuming, and difficult to scale across environments and tasks. These methods do not address the core challenge Phantom tackles: enabling true scalability by removing the need for robot data entirely.

We acknowledge that large-scale efforts to collect robot demonstrations are ongoing, and such robot data will remain very useful for co-training with human data. However, many methods that claim to scale with human videos still require a substantial amount of robot data per task to work effectively (e.g., MotionTracks \cite{ren2025motion} requires at least 22\% of its data to be teleoperated robot demonstrations; Track2Act \cite{bharadhwaj2025track2act} uses roughly 40 robot demos per task; EgoMimic \cite{kareer2024egomimic} trains on datasets where, on average, 41\% of the data comes from robot teleoperation). These methods fail in the target regime where human data is far more abundant than robot data ($\ll1\%$ robot data).

% These methods fail in the target setting where useable human data grows rapidly while robot data remains scarce.

% the target regime where human data outnumbers robot data by $100\times
% $ or more.

% he asymptotic setting where human data grows rapidly while robot data remains scarce.

In contrast, Phantom is designed to function with zero robot data, and therefore is well-positioned to succeed even as human datasets scale rapidly and robot data remains limited. While we do not demonstrate large-scale deployment in this work, our experiments provide early evidence that Phantom can operate effectively under these future data conditions. Because existing methods rely on robot data, they are unable to address Phantom's central goal — true scalability through human videos — and are thus not suitable baselines for evaluating our approach.

% For these reasons, we believe it is not meaningful to compare Phantom to methods that are constrained by their dependence on robot data since they do not address the same research objective as Phantom.

% In contrast, we are interested in the very conceivable scenario where we have at least 100 times more human data than robot data. Existing methods do not work in this $<1\%$ robot data regime.

% Human video demonstrations are significantly easier to scale than robot data, and it's conceivable that 

% In contrast, scaling human video demonstrations is si

% In contrast, human video demonstrations can scale significantly faster than robot data collection efforts. While large scale human video demonstration efforts are not publicly available yet, it is conceivable that this data will at least be 100x 

% it will at least scale to be 100x more robot data. There is no evidence to believe it would work. 

% In contrast, we are interested in a regime where human video data 

% In contrast, we aim to build a method that can operate in a regime where we have more than 10,000x more human video data than robot data. This corresponds to a regime where less than 0.01\% of the dataset is robot data. Existing methods simply do not work in this regime.

% As the size of usable human video data grows, the proportion of robot data in our training datasets will decrease. It is unclear whether methods that rely heavily on robot data will continue to perform well in this regime.

\item \textbf{The object-centric bottleneck}: Many learning from human videos methods bridge the embodiment gap using an object centric approach. However, object tracking remains an ongoing, challenging research problem, especially for multi-object tracking, deformables, granular materials, and liquids. As a result, these methods fail on these types of objects. Phantom has no such constraints and has been shown to work with groups of objects and deformable objects. A fair comparison to Phantom should include works that are not specialized for one type of object.
\end{enumerate}

\textbf{EgoMimic} \cite{kareer2024egomimic} proposes a method to learn robot policies from human videos. EgoMimic and Phantom differ in two main ways:
\begin{itemize}
    % \item \textbf{Hardware}: EgoMimic relies on Project Aria glasses for their hand pose estimation. At the time of writing, these glasses are not commercially available. Phantom requires only an RGBD camera. As a result, comparing the quality of our hand pose estimation to EgoMimic's is infeasible.
    \item \textbf{Strategy to bridge visual embodiment gap}: EgoMimic attempts to bridge the visual embodiment gap using the Red Line overlay whereas Phantom uses a virtual robot overlay. Our experiments convincingly show that the virtual robot overlay works significantly better. EgoMimic only shows results for a much easier cross-embodiment setup where the robot arm is already closely aligned with the human arm, and it is possible that in this easier setup the EgoMimic data editing strategy would have more success than in our more challenging scenario. 
    \item \textbf{Actions}: EgoMimic leverages human data only to improve predictions of the gripper's pose, and relies entirely on robot data to predict the width of the gripper. As a result, EgoMimic requires at least two hours of robot data collection before being able to cotrain with human data. A full comparison to EgoMimic would require adapting our Red Line overlay baseline to rely on robot data to determine the opening width of the gripper. Doing so would defeat the purpose of evaluating how well our method transfers policies from only human videos to robots, since learning gripper opening widths from robot data does not achieve the goal of transferring policies exclusively from humans to robots.

    % Due to the hardware constraints outlined above, we cannot directly compare with EgoMimic's hand pose estimation method. Therefore, the best comparison we can run is to use EgoMimic's visual data editing approach with our hand pose estimation method that predicts the full robot pose (which corresponds to our Red Line baseline). An even closer comparison to EgoMimic would be to use our hand pose estimation only to extract the pose of the gripper and rely exclusively on robot data to predict the opening width of the gripper, as done in EgoMimic. However, doing so would defeat the purpose of evaluating how well our method transfers policies from only human videos to robots, since learning gripper opening widths from robot data does not achieve the goal of transferring policies exclusively from humans to robots. 
\end{itemize}

\textbf{MotionTracks} \cite{ren2025motion} proposes a unified image-space action representation—2D motion tracks—for imitation learning from both human videos and robot demonstrations. However, all experiments include datasets with at least 22\% teleoperated robot data. In contrast, our work focuses on building a highly scalable framework that learns exclusively from human videos, avoiding the teleoperation bottleneck entirely. By removing this bottleneck, Phantom tackles a harder problem with greater potential to scale. 

\textbf{Track2Act} \cite{bharadhwaj2025track2act} learns 2D point tracks from large-scale web videos and lifts them to 3D object transformations to obtain robot actions, which are then refined by a residual closed-loop policy trained on teleoperated demonstrations. Despite leveraging 400,000 video clips, Track2Act still requires around 40 robot demos per task. This reliance on robot data remains a major bottleneck to scaling— the challenge Phantom aims to overcome. Phantom learns policies without any robot data, enabling scalable learning directly from human videos. Relying on co-training with robot data, as in Track2Act, does not address our core research objective.

\textbf{Im2Flow2Act} \cite{xu2024flow} uses object flow to transfer human video demonstrations to real-world robot manipulation without any real robot training data. It learns policies from simulated object flow, requiring a manually constructed simulation environment and task-specific heuristic actions. This involves substantial overhead, as humans must create simulation environments with relevant, similar objects and define heuristics for each task. While simulation-based transfer is a valuable direction, our work focuses on policy transfer without the manual effort of constructing simulation environments or hand-designing task heuristics.

\textbf{R+X} \cite{papagiannis2024r+} automatically retrieves human videos that are relevant for a target task and executes a robot policy conditioned on keypoint trajectories from these human videos. However, R+X is only demonstrated to operate in an open-loop manner. In contrast, we enable closed-loop execution, which is critical for robustness—especially in tasks like Sweep Trash, where the robot must adapt to unpredictable object dynamics.

\textbf{Orion} \cite{zhu2024vision} enables robots to learn manipulation tasks from a single human video in open-world settings using object-centric reasoning via Open-world Object Graphs (OOGs). It extracts keyframes and models object relations and motions to learn a policy. However, Orion relies on computing the desired SE(3) transformation of the target object to derive robot actions—a strategy that breaks down for deformable objects and multiple objects. In contrast, Phantom imposes no such constraint; for example, we demonstrate tying a figure-eight knot with a rope. 

\textbf{Vid2Robot} \cite{jain2024vid2robot} learns an end-to-end video-conditioned policy from paired videos of humans and robots performing the same task, enabling alignment across embodiments and environments. However, the need for such paired data significantly limits scalability, and therefore does not address our core research objective. 

\textbf{HOPMan} \cite{bharadhwaj2024towards} predicts future hand-object interaction plans from passive human videos using a diffusion model, and then translates these plans into robot actions via a policy trained on paired human-robot data. However, this approach required 3 days of data collection to obtain 600 paired demonstrations, plus an additional 1,000 robot-only demos to deploy across 16 skills. While promising, this heavy reliance on robot data remains a major bottleneck to scalability.

\textbf{DITTO} \cite{heppert2024ditto} is a two-stage framework that extracts object-centric trajectories by segmenting and tracking objects offline, then re-detects and re-localizes these objects at deployment to warp the demonstration trajectory accordingly. It executes the warped trajectory using an off-the-shelf grasp planner. However, DITTO relies on estimating the desired SE(3) motion of the target object, which fails for deformable objects and multiple objects—a limitation Phantom avoids entirely. Moreover, DITTO does not report quantitative results for its real-world experiments.

\textbf{MimicPlay} \cite{wang2023mimicplay} is a hierarchical imitation learning framework that learns high-level task plans from unlabeled human play videos and uses them to guide low-level robot control trained on teleoperated robot demonstrations. EgoMimic \cite{kareer2024egomimic} has already been shown to outperform MimicPlay, and we compare to EgoMimic to the extent possible. Additionally, MimicPlay is trained on a dataset in which at least 75\% of the data consists of robot demonstrations, as measured by data collection time. MimicPlay's heavy reliance on robot data makes it unsuitable for comparison.

\textbf{AR2-D2} \cite{duan2023ar2} is a system for collecting robot manipulation demonstrations without requiring a physical robot, using an iOS app that overlays a virtual robot onto real-world scenes and tracks the user’s hand to generate robot-aligned trajectories. However, this method requires that videos be collected with an iPhone/iPad and that the human video demonstrator manually annotate keypoint poses for the robot to follow live during data collection. This makes collecting human video demonstrations significantly slower and constrains the type of motions that can be demonstrated, as evidenced by the very limited tasks shown in the paper: pick (without placing), press, and push. 

\textbf{WHIRL} \cite{bahl2022human} learns policies by extracting interaction priors from third-person human videos and refining them through real-world robot interactions using sampling-based policy optimization. However, its reliance on physical robot trials limits scalability. In contrast, Phantom directly produces high-quality actions from human videos without requiring any robot interaction beforehand.

\textbf{Learning by Watching} \cite{xiong2021learning} uses an image-to-image translation network to map a single human video into the robot domain, extracts keypoints from the translated images, and builds a reward function to train a policy in simulation. However, this method is only demonstrated to work in a single scene in simulation, while our work focuses on real-world deployment.

\begin{table}[H]
\centering
\begin{tabular}{lccc}
\toprule
 & \makecell{No Robot \\Data} & \makecell{Deformable \\Objects} & \makecell{Closed-loop} \\
\cmidrule{1-4}
EgoMimic \cite{kareer2024egomimic} & \redcross & \greencheck & \greencheck \\
MotionTracks \cite{ren2025motion} & \redcross & \greencheck & \greencheck \\
Track2Act \cite{bharadhwaj2025track2act} & \redcross & \redcross & \greencheck \\
Im2Flow2Act \cite{xu2024flow} & \redcross$^*$ & \greencheck & \greencheck \\
R+x \cite{papagiannis2024r+} & \greencheck & \greencheck & \redcross \\
ORION \cite{zhu2024vision}& \greencheck & \redcross & \greencheck \\
Vid2Robot \cite{jain2024vid2robot} & \redcross & \greencheck & \greencheck \\
HOPMan \cite{bharadhwaj2024towards} & \redcross & \greencheck & \greencheck \\
DITTO \cite{heppert2024ditto} & \greencheck & \redcross & \greencheck \\
Mimicplay \cite{wang2023mimicplay} & \redcross & \greencheck & \greencheck \\
AR2-D2 \cite{duan2023ar2} & \redcross & \greencheck & \greencheck \\
WHIRL \cite{bahl2022human} & \redcross & \greencheck & \greencheck  \\
Learning by watching \cite{xiong2021learning} & \redcross$^*$ & \redcross & \greencheck \\
\rowcolor{yellow!30} \textbf{Ours }& \greencheck & \greencheck & \greencheck \\
\bottomrule
\end{tabular}
\vspace{0.2cm}
\caption{Comparison between our method and other related works. \textbf{No Robot Data}: the method does not require robot data in policy training. \redcross$^*$ indicates that the method relies on simulation data which is limited by the need to create simulation environments that are representative of real world interactions. \textbf{Deformable Objects}: the method is demonstrated to work on deformable objects. \textbf{Closed-loop}: the method is closed-loop. }
\label{table:related_works_comparison}
\end{table}

\clearpage
\subsection{Phantom is robot agnostic}

\begin{figure}[h]
    \centering
    \includegraphics[width=1.0\linewidth]{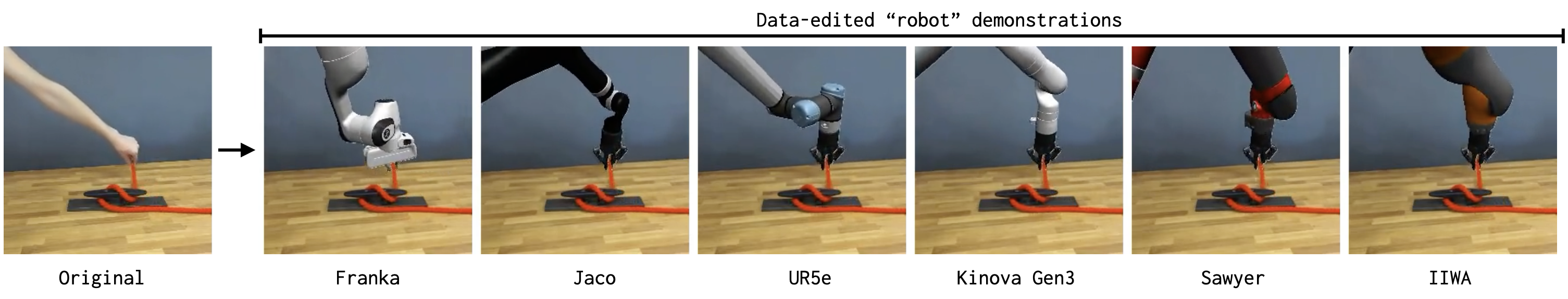}
    \caption{Our method is robot agnostic. Each human video can be converted into a robot demonstration for any robot capable of completing the task.}
    \label{fig:multi_robots}
\end{figure}

\vspace{1em}
\subsection{Evaluating the Benefits of Co-training with Diverse Human Data}

While we have already shown in previous experiments that our policy can be deployed zero-shot without any robot data, we also investigate the benefits of co-training with robot data given that there already exists considerable amounts of robot data. To do this, we collect 100 teleoperated demonstrations in a single scene on the Kinova robot using an Oculus controller for the sweeping task. In each observation image from these demonstrations, we overlay a virtual Kinova on the real robot to align the visual distributions between our datasets. Next we co-train our robot data with our larger scale human video dataset consisting of 950 demonstrations in many different scenes. We see that while the robot-only policy performs well in-distribution, its success rate drops to zero in a new scene. Co-training with human videos from diverse scenes, however, increases the performance to 80\%. 

\begin{table}[H]
\centering
\begin{tabular}{lccc}
\toprule
               & \makecell{In-distribution \\ Scene} & \makecell{Out-of-distribution \\ Scene} \\
\midrule
Robot     & 0.88  & 0.0 \\
Robot + Human   & \textemdash  & 0.80 \\

\bottomrule
\end{tabular}
\vspace{0.3cm}
\caption{\textbf{Co-training with diverse human data}. Evaluating the benefits of co-training with diverse human data. 25 rollouts per evaluation.}
\label{table:ablation}
\end{table}

\subsection{Comparing Human vs. Robot Data}
While our approach significantly reduces the cost of scaling data collection across diverse environments, it introduces a tradeoff: human video demonstrations provide scalability at the expense of some precision, due to uncertainty in hand pose estimation from RGBD videos. To investigate how much precision is lost, we compare policies trained on teleoperated robot demonstrations (collected using an Oculus controller) against those trained on human video demonstrations for the Kinova sweeping task. All data is collected and evaluated in the same scene.

As shown in Table~\ref{table:robot_vs_human}, a policy trained on 50 teleoperated demonstrations achieves a 52\% success rate, compared to 44\% for 50 human demonstrations—a modest drop that is not statistically significant ($p = 0.778$, where significance is defined as $p < 0.05$). At 100 demonstrations, the gap widens (88\% vs. 64\%), but still falls short of statistical significance ($p = 0.095$). Scaling to 300 human demonstrations improves success to 84\%. These results suggest that while individual human demonstrations may be less precise, their ease of collection allows for scaling that effectively closes the performance gap.

\begin{table}[H]
\centering
\vspace{-0.5cm}
\begin{tabular}{l|ccc}
\toprule
\# demos & Robot only & Human only & p-value (Fisher) \\
\midrule
50 & 0.52 & 0.44 & 0.778 \\
100 & 0.88 & 0.64 & 0.095 \\
300 & \textemdash & 0.84 & \textemdash \\
\bottomrule
\end{tabular}
\vspace{0.4cm}
\caption{\textbf{Performance of policies trained on robot vs. human demonstrations.} 
Each configuration was evaluated with 25 rollouts. P-values (Fisher’s exact test) assess whether success rates differ significantly ($p < 0.05$) between robot and human data. While human only policies exhibit a drop in performance, no statistically significant differences were found. The robot-only condition for 300 demos was not trained due to the significant time requirement and because it would not meaningfully add to this analysis.}
\label{table:robot_vs_human}
\end{table}

Importantly, this single scene comparison understates the strength of our approach. \textbf{The central value of human video data is its scalability across environments.} Collecting robot data in many scenes requires moving physical hardware—an expensive and time-consuming process—whereas human video data can be gathered quickly and at scale with minimal overhead. A full comparison should also compare the time required to collect diverse human data across many scenes and the time required to collect equivalent robot data across those same scenes. However, we do not perform this experiment because robot data collection in many scenes is prohibitively slow.

\subsection{Data collection Details}
\label{section:appendix-dataset}
The details of the human video demonstration datasets collected for all tasks are outlined in Table \ref{table:dataset_detail}. 

\begin{table}[H]
\centering
\begin{tabular}{lcccc}
\toprule
\multirow{6}{*}{\vspace{-2.0em} Panda tasks} & Task & Number of demos & Max Horizon\\
\cmidrule{1-4}
&Zebra & 313 & 228\\
&Stack & 268 & 148\\
&Sweep & 279 & 407\\
&Cleat & 307 & 516\\
&Soap & 283 & 175\\
\cmidrule{1-4}
Kinova task & Sweep & 950 & 168\\
\bottomrule
\end{tabular}
\vspace{0.3cm}
\caption{Human video datasets. \textbf{Number of demos}: number of human video demonstrations used in training, \textbf{Max Horizon}: the maximum number of steps in the human video demonstration dataset. }
\label{table:dataset_detail}
\end{table}

\begin{figure}[H]
    \centering
    \includegraphics[width=0.7\linewidth]{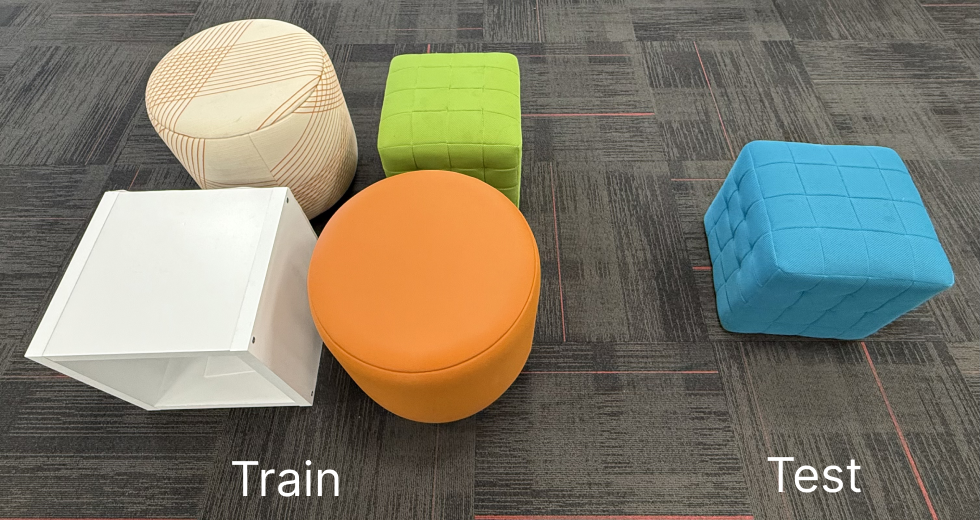}
    \caption{Left: surfaces used during human video data collection for the kinova sweep task. Right: Unseen surface used to evaluate policy.}
    \label{fig:surfaces}
\end{figure}

\subsection{Policy Training Details}
Table \ref{table:training_hyperparameters} describes the hyperparameters used to train diffusion policy. All experiments done on the Kinova robot use the same set of hyperparameters. This includes the diverse scene experiments, inpainting quality experiments, and robot vs. human video data comparison experiments. We used the same set of hyperparameters for all data-editing methods across each task. All tasks in the paper use the DDIM scheduler with 100 training steps and 10 inference steps.

\begin{table}[!htbp]
\centering
\small
\setlength{\tabcolsep}{3.5pt} % Reduce column spacing
\renewcommand{\arraystretch}{0.9} % Reduce row height
\begin{tabular}{lccccccc}
\toprule
\makecell{Robot} & Task & To & Ta & ImgRes & Batch & Lr & Aug\\
\midrule
\multirow{5}{*}{\makecell{Panda}} 
& Zebra & 2 & 8 & 240 & 200 & 1e-4 & Yes\\
& Stack & 2 & 8 & 240 & 200 & 1e-4 & Yes\\
& Sweep & 2 & 8 & 240 & 200 & 1e-4 & Yes\\
& Cleat & 2 & 8 & 240 & 200 & 1e-4 & Yes$^*$\\
& Soap & 2 & 8 & 240 & 200 & 1e-4 & Yes\\
\midrule
\makecell{Kinova} & Sweep & 2 & 8 & 240 & 256 & 1e-4 & Yes\\
\bottomrule
\end{tabular}
\vspace{0.3cm}
\caption{Hyperparameters for Diffusion policy training. \textbf{To}: observation horizon, \textbf{Ta}: action horizon. \textbf{Aug}: Image augmentations (RandomCrop, RandomRotation, ColorJitter) used during training. Only ColorJitter was used for the Cleat task to avoid cropping out grasps of the rope that frequently occur on the edge of the image.}
\label{table:training_hyperparameters}
\end{table}

All virtual robot overlays are generated using models from Mujoco Menagerie \cite{menagerie2022github}.

\subsection{Detailed Task Descriptions}
\begin{figure}[h]
    \centering
    \includegraphics[width=0.9\linewidth]{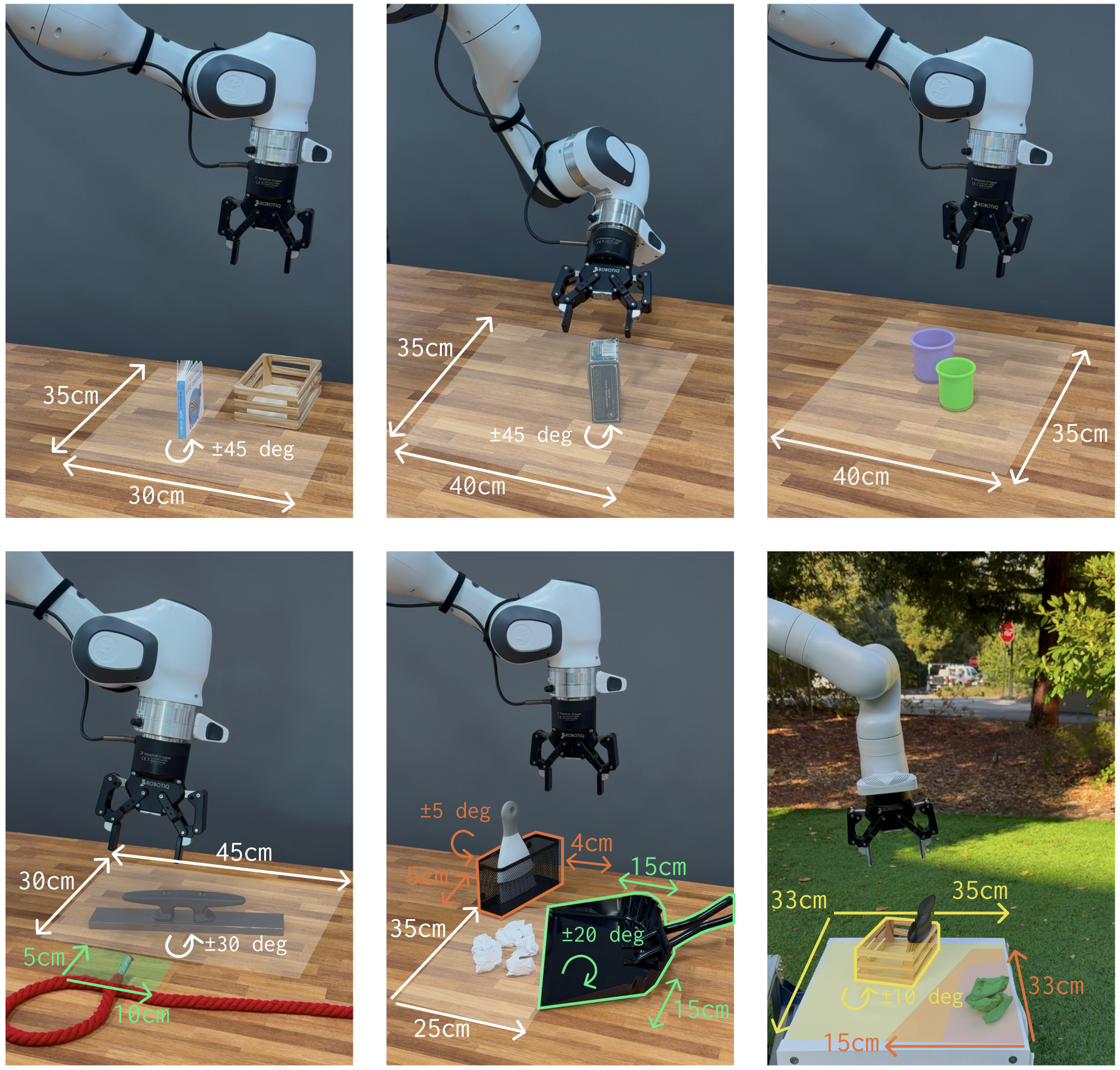}
    \caption{Variation in object placement during evaluations of each task.}
    \label{fig:task_description}
\end{figure}

The variation in the placement of objects for each task is visualized in Fig. \ref{fig:task_description}.

\textbf{Pick and Place Book}: The robot must pick up a book and place it inside a wooden container. The book's initial position is randomly sampled within the outlined 30 cm × 35 cm rectangular region. Additionally, its orientation can vary by ±45 degrees.

\textbf{Rotate Box}: The robot must rotate a box 90 degrees onto a new face in a controlled fashion (simply knocking it over is not valid). The box's initial position is randomly sampled within the outlined 40 cm x 35 cm rectangular region.

\textbf{Stack Cups}: The robot must stack the green cup inside the purple cup. Precise alignment is critical, as the cups differ in diameter by only 1.5 cm. The cups' initial positions are randomly sampled within the outlined 40 cm x 35 cm rectangular region.

\textbf{Tie Rope}: The robot must tie a simplified cleat hitch, a sailing knot that follows a figure-eight $\infty$ pattern. This task is challenging due to the precise manipulation required of a highly deformable object. The position of the cleat is randomly sampled within the outlined 30cm x 45cm white rectangular region and rotated by ±30 degrees.

\textbf{Franka Sweep Trash}: The robot must pick up the sweeper and sweep six pieces of trash into a dustpan. This task involves coordinated multi-object manipulation, requiring the robot to control the sweeper while simultaneously managing the movement of multiple loose objects. Additionally, the pieces of trash exhibit unpredictable dynamics, necessitating continuous adaptation based on real-time feedback. The position of the six pieces of trash is randomly sampled within the 25cm x 35cm rectangular region outlined in white. The position of the sweeper is randomly varied by 5cm x 4cm laterally and rotated by ±5 degrees. The position of the dustpan is randomly varied by 15cm x 15cm laterally and rotated by ±20 degrees.

\textbf{Kinova Sweep Trash}: The robot must pick up the sweeper and sweep the green piece of trash off the table. The position of the sweeper is randomly sampled within the 33cm x 35cm rectangular region outlined in yellow. The position of the green piece of trash is randomly sampled within the 15cm x 33cm region outlined in orange.

 \subsection{Hand Mask Data Editing Method}
\label{section:shadow_details}
We describe in detail how we adapt the data-editing strategy from Shadow \cite{lepertshadow} to the human-to-robot setting. 

\textbf{Data-editing during training:} We segment out the pixels corresponding to the hand and set them to black. We extract the hand pose using the strategy described in Section \ref{section:action_labels} and overlay an RGB rendering of the target robot in this pose using the known camera extrinsics.

\textbf{Data-editing at inference time:} To ensure that the train and test time images match closely, we overlay a black segmentation mask of the human arm and hand in the same pose as the robot. However, unlike in Shadow where the authors transferred policies between two robots, we do not have access to a realistic virtual model of a human arm and hand. Instead we train a diffusion model to predict the segmentation mask of the hand given a 6-DOF pose. This diffusion model is trained from scratch using the hand masks and corresponding target poses from our training dataset. At test time, we overlay the segmentation mask predicted by the diffusion model onto our image, and feed this edited image into our policy.

\subsubsection{Training the Hand Mask Diffusion Model}
We train the diffusion model for generating hand masks using the hyperparameters in Table \ref{table:training_hyperparameters_handdiff}. Due to high compute requirements, we generate 64×64 images and upscale them to the desired resolution using a super-resolution model \cite{Lim_2017_CVPR_Workshops}. To improve temporal consistency, the Hand Mask Diffusion Model generates hand masks for both time $t$ and $t+1$ given a robot pose at time $t$. Additionally, we apply attention injection \cite{Tumanyan_2023_CVPR} at test time, using the model’s output at time $t$ as the reference image for time $t+1$.

\begin{table}[!htbp]
\centering
\small
\setlength{\tabcolsep}{4pt} % Reduce column spacing
\renewcommand{\arraystretch}{0.9} % Reduce row height

\begin{tabular}{lccccc}
\toprule

\makecell{Forward Diffusion \\ Timesteps} & \makecell{Sampling \\ Steps} & ImgRes & Batch & Lr \\
\midrule
 1000 & 50 & 64 & 32 & 1e-4 \\
\bottomrule
\end{tabular}
\vspace{0.2cm}
\caption{Hyperparameters used to train the Hand Mask diffusion model. \textbf{Forward Diffusion Timesteps}: the number of forward process steps at train time. \textbf{Sampling Steps}: the number of sampling steps used during inference.}
\label{table:training_hyperparameters_handdiff}
\end{table}

\subsubsection{Data Augmentation}
Because the angle of the human arm can vary for the same 6-DOF pose, we need to make our policy agnostic to this variation. In addition, we find that the diffusion model we use to render the human segmentation mask does not output temporally consistent angles of the arm across frames. To address this problem, we augment our training data to include a second segmentation mask of the arm that is randomly shifted by a few pixels at each timestep. Importantly, the segmentation mask of the robot is not shifted, ensuring that the model can rely on the segmentation mask of the robot to localize the embodiment without relying on the hand mask.

We evaluate the impact of this augmentation on the Pick and Place Book and Stack Cups tasks. As shown in Table \ref{table:hand_mask_aug_ablation}, incorporating the augmented hand mask achieves comparable performance in the easier Pick and Place Book task, but increases the success rate for the harder Stack Cups task by 12 percentage points. This suggests that the augmentation helps mitigate inconsistencies in the human arm mask generation, leading to more robust policy learning.

\begin{table}[H]
\centering
\begin{tabular}{lccc}
\toprule
& \makecell{Pick/ Place Book} & \makecell{Stack Cups} \\
\midrule
Hand Mask & 0.92 & 0.52 \\
Hand Mask (no data aug) & 0.88 & 0.40\\
\bottomrule
\end{tabular}
\vspace{0.3cm}
\caption{\textbf{Effect of data augmentation on the Hand Mask strategy.} Adding random shifts to the hand mask during training improves performance in Stack Cups, where success increases from 40\% to 52\%. This suggests that the augmentation helps the policy generalize despite inconsistencies in the human arm segmentation. 25 rollouts per evaluation.}
\label{table:hand_mask_aug_ablation}
\end{table}